\documentclass[runningheads]{llncs}
\usepackage[T1]{fontenc}
\usepackage{graphicx}
\usepackage{hyperref}
\usepackage{multirow}
\usepackage{amsmath}
\usepackage{amssymb}
\usepackage{comment}
\usepackage{booktabs}

\begin{document}
%
\title{Schrödinger Diffusion Driven Signal Recovery in 3T BOLD fMRI Using Unmatched 7T Observations}
\titlerunning{Enhancing 3T fMRI SNR with unpaired 7T Data}
%
%

\author{Yujian Xiong\inst{1} \and Xuanzhao Dong\inst{1} \and Sebastian Waz\inst{3} \and Wenhui Zhu\inst{1} \and Negar Mallak\inst{1} \and
Zhong-Lin Lu\inst{2,3,4,*} \and Yalin Wang\inst{1,*}}
\authorrunning{Y. Xiong et al.}

%
\institute{School of Computing and Augmented Intelligence, Arizona State University, Tempe, AZ, USA \and
Division of Arts and Sciences, New York University Shanghai, Shanghai, China \and 
Center for Neural Science and Department of Psychology, New York University, New York, NY, USA \and
NYU-ECNU Institute of Brain and Cognitive Science, NYU Shanghai, Shanghai, China}
\maketitle              
\begin{abstract}
Ultra-high-field (7 Tesla) BOLD fMRI offers exceptional detail in both spatial and temporal domains, along with robust signal-to-noise characteristics, making it a powerful modality for studying visual information processing in the brain. However, due to the limited accessibility of 7T scanners, the majority of neuroimaging studies are still conducted using 3T systems, which inherently suffer from reduced fidelity in both resolution and SNR. To mitigate this limitation, we introduce a new computational approach designed to enhance the quality of 3T BOLD fMRI acquisitions. Specifically, we project both 3T and 7T datasets, sourced from different individuals and experimental setups, into a shared low-dimensional representation space. Within this space, we employ a lightweight, unsupervised Schrödinger Bridge framework to infer a high-SNR, high-resolution counterpart of the 3T data, without relying on paired supervision. This methodology is evaluated across multiple fMRI retinotopy datasets, including synthetically generated samples, and demonstrates a marked improvement in the reliability and fit of population receptive field (pRF) models applied to the enhanced 3T outputs. Our findings suggest that it is feasible to computationally approximate 7T-level quality from standard 3T acquisitions.

\end{abstract}
\section{Introduction}

To unravel the complex mechanisms underlying visual encoding and decoding in the human brain, investigators have utilized diverse methodologies to model both the spatial and temporal variation of brain activity, often measured as blood-oxygenation-level-dependent (BOLD) signals collected via functional magnetic resonance imaging (fMRI). Many computational approaches to modeling visually evoked BOLD signals focus on independently modeling the time series at each voxel, e.g., phase-dependent models \cite{engel1994fmri,deyoe1996mapping} used to measure visual eccentricity, the now standard population receptive field (pRF) model introduced by Dumoulin and Wandell~\cite{dumoulin2008population} then developed by others \cite{kay2013compressive}, and more sophisticated deep learning techniques \cite{thielen2019deeprf}. By modeling the voxels within a given cortical region in this way, researchers are able to delineate retinotopic maps, topology-preserving representations of the visual field on the brain's surface \cite{wandell2007visual}. Although the existence of retinotopic maps has been known for over a century~\cite{ribeiro2024human}, the possibility of mapping them out non-invasively using fMRI is a relatively recent development \cite{dumoulin2003automatic}. Today, retinotopic mapping shows promise as a clinical endpoint, e.g., in tracking glaucoma \cite{duncan2007retinotopic} and other neurodegenerative conditions like Alzheimer's disease \cite{brewer2014visual}.

A current challenge in retinotopic mapping is limited access to high-quality fMRI scans. Although the current largest freely available retinotopy datasets, such as the Human Connectome Project~\cite{uugurbil2013pushing,van2013wu} and Natural Scenes Dataset~\cite{allen2022massive}, was collected using a 7-Tesla (7T) fMRI machine capable of relatively high-resolution and high signal-to-noise ratio (SNR) imaging, this enhanced resolution is not concentrated in the occipital lobe, where retinotopic maps are most prominently studied. Still, these data provide superior resolution and SNR when compared to measurements from more widely available 3-Tesla (3T) machines which were used to generate similar datasets for more general tasks~\cite{chang2019bold5000,horikawa2017generic,gong2023large}.


Access to such high-quality fMRI data would have an immediate benefit within retinotopic mapping and beyond. For example, a number of standard atlases\cite{wang2015probabilistic,glasser2016multi} used to identify brain areas such as retinotopic maps are, effectively, averages over many subjects and are thus limited in resolution by their underlying data. Poor resolution can also pose an obstacle to pRF modeling of the time series at each cortical vertex, since the formation of these time series from raw fMRI often involves some spatial smoothing \cite{glasserSupplementaryMaterialMultimodal2016}. Retinotopic mapping is often performed on top of pRF model solutions \cite{dougherty2003visual,benson2022variability,himmelberg2023comparing}, but pRF solutions formed from low-quality data may lead to topological violations in the map which disagree with knowledge of cortical physiology and must be corrected \cite{tu2021topology,xiong2023characterizing}.

With the success of deep learning models in computer vision, researchers have extended these techniques to medical imaging. Generative Adversarial Networks (GANs)~\cite{goodfellow2020generative} and their variants~\cite{armanious2019unsupervised,phan2023structure,zhu2023otre}, as well as conditional diffusion models~\cite{ho2020denoising,sasaki2021unit,konz2024anatomically,dong2024cunsb,chen2024contourdiff}, have shown promising results. However, many of these methods rely on paired data or struggle with unpaired domain alignment, limiting their applicability to certain medical imaging tasks. On the other hand, despite their broad application on various modalities, including fundus imaging~\cite{dong2024cunsb,shen2020modeling}, MRI-CT~\cite{li2020magnetic,zhu2024diffusion,cui20247t,huang2024noise,zhang2024phy}, and fMRI for natural image reconstruction~\cite{fang2020reconstructing,ren2021reconstructing,takagi2023high,chen2024cinematic}, there has been limited focus on enhancing fMRI signals to improve SNR and retinotopic mappings. This gap highlights the need for methods that specifically target the enhancement of fMRI signals and downstream retinotopic analysis.

To address these limitations, we propose a novel framework that enhances these scans using unsupervised learning. Our approach maps 3D brain surfaces into a shared planar parametric domain via conformal mapping and applies an unpaired Schrödinger Bridge~\cite{leonard2013survey,kim2023unpaired,dong2024cunsb,wang2024implicit} diffusion model to enhance 3T BOLD signals. The resulting fMRI data preserve structural integrity while approximating the quality and distribution of high-resolution 7T scans, overcoming challenges posed by short-duration and low-resolution 3T fMRI experiments.

In summary, our study introduces a novel approach to fMRI enhancement with the following contributions: (a) We develop a robust fMRI enhancement pipeline that starts from raw fMRI data across different subjects and datasets. (b) To the best of our knowledge, it is the first work who addressed fMRI and retinotopic mapping through unpaired diffusion model. (c) We validate our framework on real and synthetic datasets, demonstrating its capability to produce high-quality fMRI scans and may improve various downstream such as retinotopic mapping through pRF analysis.



\section{Methods}

\begin{figure}[t]
    \centering
    \includegraphics[width=0.9\linewidth]{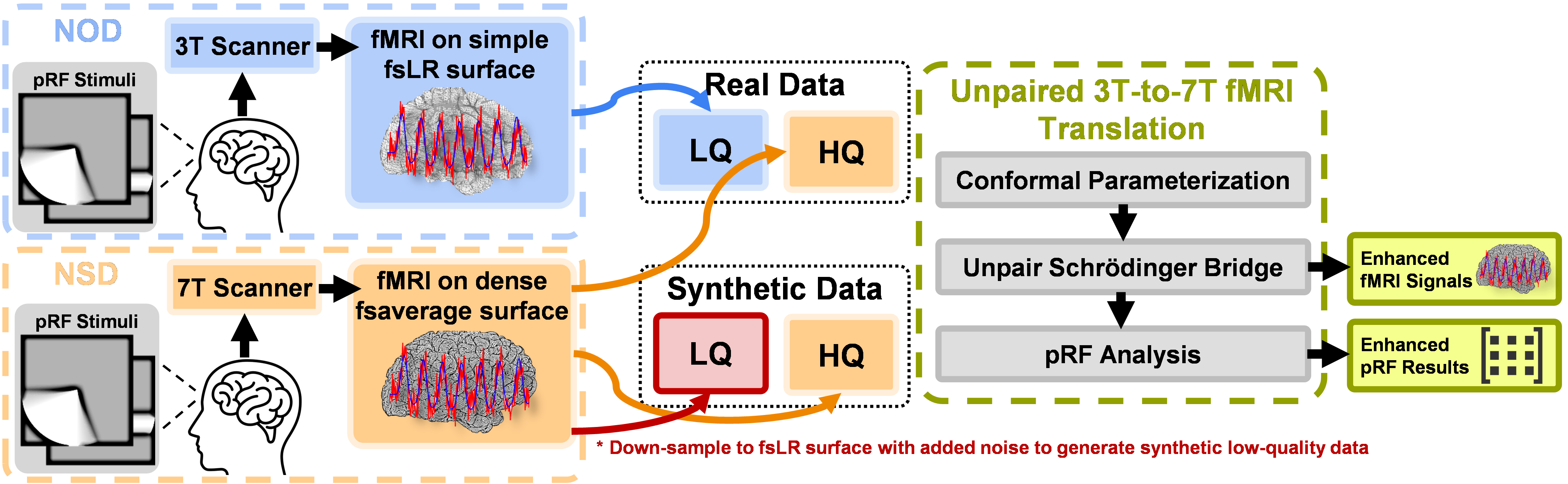}
    \caption{Overview of the Pipeline: We collect 3T and 7T fMRI data from the NOD and NSD datasets, respectively. The fMRI signals at each time step from all pRF experiments are used to generate low-quality (LQ) and high-quality (HQ) fMRI slices. For synthetic data, the original NSD fMRI serves as the ground truth, while their down-sampled versions as LQ data. All data are fed into our pipeline with three components.}
    \label{fig:pipeline}
\end{figure}

Fig.~\ref{fig:pipeline}, illustrate our pipeline of the unpaired Schrödinger Bridge enhancement.



\subsection{Datasets and Experimental Designs}\label{sec:Method-data}

We utilize two datasets: the 7T \textit{Natural Scenes Dataset} (NSD)~\cite{allen2022massive} and the 3T \textit{Natural Object Dataset} (NOD)~\cite{gong2023large}. The NSD contains approximately 40 sessions per subject for 8 participants, including natural scene images and pRF-fLoc stimuli~\cite{benson2018human,stigliani2015temporal}, providing high-quality, fine-resolution fMRI data. In contrast, the NOD includes 10 to 63 sessions per subject for 30 participants, with 9 subjects performing pRF-fLoc tasks and the rest viewing only images, offering broader but lower-quality 3T scans in similar retinotopic experiments.

Due to the limitations of the magnetic field strength, 3T datasets suffer from reduced spatial resolution, temporal resolution, and signal-to-noise ratio (SNR). Conversely, 7T datasets offer superior resolution on fine-grained brain structures and significantly higher SNR. We leverage the 7T NSD data as the target for our enhancement model, while the 3T NOD data serves as the source. We extract time-step slices to create unpaired low-quality and high-quality fMRI collections, as shown in Fig.~\ref{fig:pipeline}.

For robust evaluation, we design two setups: \textbf{(a) Real Data}: All 8 NSD subjects are used as high-quality targets, while 9 NOD subjects with pRF tasks serve as low-quality sources, where $s_1 \sim s_7$ are used to train the model, and $s_8 \sim s_9$ are reserved for testing. \textbf{(b) Synthetic Data}: NSD subjects $s_1 \sim s_6$ provide high-quality data, with their down-sampled versions $s'_1 \sim s'_6$ as low-quality inputs (see details in Sec.~\ref{results:downsample}). Subjects $s_7 \sim s_8$ are reserved for testing.

A summary of the splitting strategy is shown in Tab.~\ref{t1}.



\setlength{\tabcolsep}{4pt}
\begin{table}[t]
\centering
\caption{Collection of NSD and NOD subjects in our train-test splitting strategy. For every subject $s_i$, all trials viewing the pRF stimuli are included. During training, $20\%$ randomly picked fMRI slices are reserved for validation.}
\resizebox{0.9\textwidth}{!}{%
\begin{tabular}{c|cc|cc}
\toprule
\multirow{2}{*}{ Split }  & \multicolumn{2}{c|}{Real data} & \multicolumn{2}{c}{Synthetic data} \\ \cline{2-5}
 & Source & Target & Source & Target \\ \hline
Train and Valid & NOD $s_1\sim s_7$ & NSD $s_1\sim s_8$ & NSD $s'_1\sim s'_6$ & NSD $s_1\sim s_6$ \\
Test & NOD $s_8\sim s_9$ & - & NSD $s'_7\sim s'_8$ & NSD $s_7\sim s_8$ \\  \bottomrule
\end{tabular}
}
\label{t1}
\end{table}

\subsection{Brain Disk Parameterization}\label{sec:Method-bd}

To translate low-quality fMRI slices into their high-quality counterparts, it is crucial to align the probability spaces of 3T and 7T fMRI trials. This task is challenging due to the inherently high SNR and substantial variations in brain structures across different subjects and datasets. These variations require a shared domain to which fMRI signals from specific subjects can be mapped. Such a shared domain enables 3T fMRI slices to achieve a comparable representation with 7T fMRI and facilitates unpaired translation training across multiple subjects. To address this, we leverage the 164k fsaverage~\cite{fischl2012freesurfer} cortical surface and employ conformal mapping to generate parameterized planar brain disks for our region of interest (ROI).

\subsubsection{Shared Surface Mesh and ROI.} 
We start with original 3D surface meshes from each dataset. The NSD provides native surface meshes (approximately 220k vertices per hemisphere) for each subject, along with transformation files in FreeSurfer format~\cite{allen2022massive,fischl2012freesurfer}. Using the NSD code packages~\cite{allen2022massive}, we retrieve the 3D coordinates of faces and vertices, along with vertex annotations, and transform them into the 164k fsaverage surface. In contrast, the NOD dataset stores its surface meshes in the 32k fsLR format using Ciftify~\cite{dickie2019ciftify}. The surface data and vertex annotations are retrieved via the Ciftify toolbox~\cite{dickie2019ciftify}. These meshes are then transformed into the 164k fsaverage surface using the Neuromaps toolbox~\cite{markello2022neuromaps} with linear approximation. However, the up-sampling process introduces significant distortions and topological violations in the 3T NOD fMRI data due to the resolution disparity between the original fsLR surface and the finer fsaverage surface. These challenges necessitate the application of an enhancement model for the 3T fMRI signals.

To streamline analysis, we define a ROI encompassing vertices labeled as \textit{lateraloccipital}, \textit{cuneus}, \textit{pericalcarine}, and \textit{lingual}, representing key cortical regions in the occipital lobe. This ROI ensures the inclusion of most primary visual cortex while significantly reducing computational overhead. By focusing on this ROI, we concentrate on the regions most relevant to visual experiments, optimizing both accuracy and efficiency.

\subsubsection{Conformal Mapping Parameterization.}

\begin{figure}[t]
  \centering
  \includegraphics[width=0.8\linewidth]{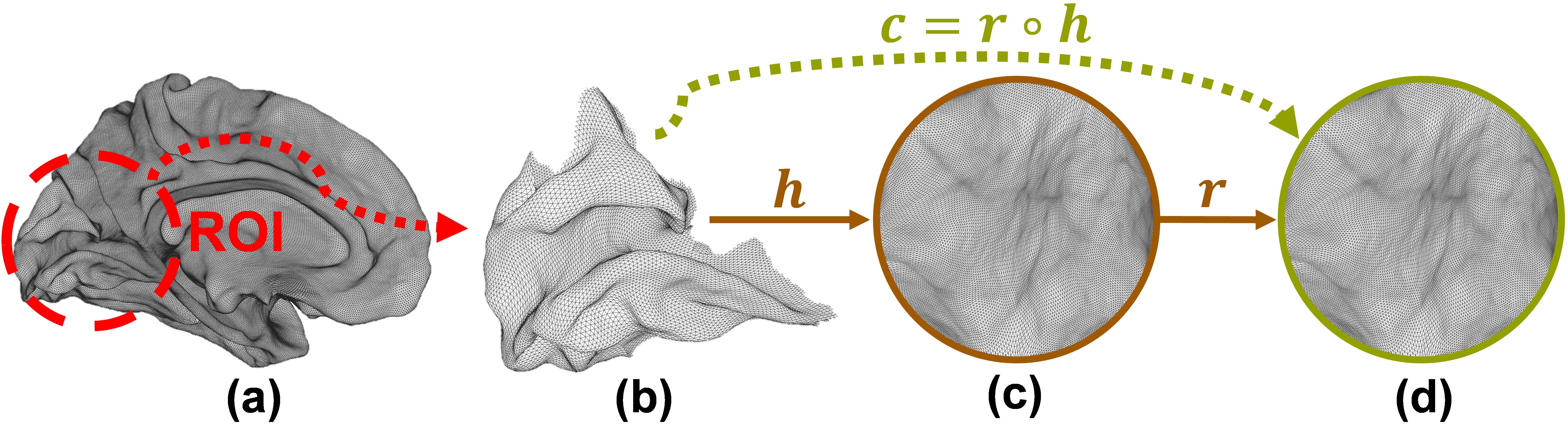}
  \caption{Disk conformal parameterization. (a) The full fsaverage cortical surface; (b) the ROI subdivision of the full mesh determined by the FreeSurfer vertex labels~\cite{fischl2012freesurfer}; (c) the parameterized planar disk of the ROI obtained from harmonic map $h$; (d) the refined planar disk through resulting conformal mapping $c = r \circ h$.}
  \label{fig:BD}
\end{figure}

In order to train our enhancement model on 2D space instead of the 3D surface meshes, we used conformal parameterization $ c: M \to D $, to map a cortical surface mesh $M$ to an unit disk $D$. Given $M$ as an open boundary genus-0 surface after cutting to ROI, the harmonic map $h: M \to D'$ minimizes the energy~\cite{jin2018conformal,gu2004genus}: $E(h) = \int_M |\nabla h|^2 \, dv_M $.

For disk-like surfaces, the harmonic map $h$ satisfies the Laplace equation~\cite{wang2007brain}: $\Delta h(u)_M = 0, \quad h|_{\partial M} = g$, where $\Delta$ denotes the Laplacian operator and $g: \partial M \to \partial D'$ is a boundary mapping given by arc-length parameterization. In discrete cases, the harmonic map is efficiently obtained by solving the sparse linear system $ L_h h = 0 $, where $ L_h $ is the Laplacian matrix~\cite{wang2007brain}.

Denote a disk to disk refinement $r : D' \to D$, the final conformal mapping would be a composition of $c = r \circ h$. To achieve conformality for mapping $c$, we can refine $r$ by iteratively modifying the mapping until its Beltrami coefficient $\mu_r$ satisfies $||\mu_r||_\infty \leq \epsilon_{\mu_r}$~\cite{wang2007brain,ta2022quantitative}.

With the conformal parameterization $c$, we can obtain a 2D parametric disk for the ROI in the 3D fsaverage mesh. This ensures a bijective and conformal parameterization for the ROI and allows us to map the fMRI signals of each vertex onto the planar disk, creating fMRI Brain Disks (BD) for every fMRI trials during the pRF experiments. The complete process is illustrated in Fig.~\ref{fig:BD}. Each visualized BD shows the instantaneous value of the BOLD signal at each vertex in the BD as RGB colors. To enhance clarity in visualization. we have omitted the edges in subsequent figures.

\subsection{Schrödinger Bridge Enhancement}\label{sec:Method-sb}

\subsubsection{Background.} The Schrödinger Bridge Problem (SBP) involves identifying the optimal stochastic process ${x_t : t \in [0, 1]}$ that transits from an initial probability distribution $p_0 \in \mathbb{R}^d$ at time $t = 0$ to a target distribution $p_1 \in \mathbb{R}^d$ at time $t = 1$. Here, $\Omega$ denotes the path space of continuous functions mapping the time interval $t \in [0, 1]$ to the state space $\mathbb{R}^d$, and $\mathcal{P}(\Omega)$ represents a potential probability measure over $\Omega$. The SBP dynamically tracks the entire process and formulates it as an optimization problem based on the difference in KL-divergence:

\begin{equation}\label{eq:sb-dynamic}
T^{\star} = \arg\min_{T \in \mathcal{Q}(p_0,p_1)} D_{\text{KL}}(T \| W^\tau) 
\end{equation}

Here, $W^\tau$ denotes the reference measure (i.e., the Wiener measure with variance $\tau$), and $\mathcal{Q}(p_0,p_1) \subset \mathcal{P}(\Omega)$ is a stochastic process that requires the boundary distributions (i.e., start at $p_0$ and end at $p_1$). The solution $T^{\star}$ is called the Schrödinger Bridge (SB), which connects the boundary distributions $(p_0, p_1)$ and works as an optimized result for the entire trajectory.


In the context of enhancing parameterized BDs, the problem is framed as a domain-transfer question,  where $p_0$ and $p_1$ represent the distributions of the fMRI response corresponding to 3T and 7T experiments, respectively. The solution $T^{\star}$ identifies the most likely evolution path of BDs over $t$, providing a probabilistic trajectory that bridges the two distributions.


\subsubsection{Discrete Bridge Approximation.}
Rather than continuously tracking the entire stochastic process, which tends to be computationally expensive, the SB described in Eq.~\ref{eq:sb-dynamic} can be approximated as a sequence of solutions to Entropic Optimal Transport (EOT) problems over successive time intervals~\cite{korotin2023light,dong2024cunsb,kim2023unpaired}. Specifically, let $[t_a, t_b] \subseteq [0, 1]$ represent an arbitrary sub-interval. The Schrödinger Bridge couplings, denoted as $T^{\star}_{t_a, t_b}$, shares the same joint marginal distribution as $T^{\star}$ over $t_a$ and $t_b$, then formulated as the solution to an EOT problem:

\begin{equation}
T^{\star}_{t_a, t_b} = \arg\min_{\gamma \in \Pi(T_{t_a}, T_{t_b})} \mathbb{E}_{(x_{t_a}, x_{t_b}) \sim \gamma} \|x_{t_a} - x_{t_b}\|^2 - 2\tau (t_b - t_a) H(\gamma)
\label{eq:SBsim1}
\end{equation}

\noindent where $\gamma \in \Pi(T_{t_a}, T_{t_b})$ represents all possible joint marginal distributions consistent with the boundary states $T_{t_a}$ and $T_{t_b}$(i.e., the most likely distributions of BDs over arbitrary time interval), and $H( \cdot )$ denotes the entropy function with hyperparameters $\tau$. Additionally, for $t \in [t_a, t_b]$, the conditional distribution $p(x_t \mid x_{t_a}, x_{t_b})$ follows the Gaussian distribution~\cite{tong2023conditional}, expressed as:
\begin{equation}\label{eq:SBsim2}
\begin{split}
&p(x_t \mid x_{t_a}, x_{t_b}) \sim \mathcal{N}(x_t; \mu_{t_a, t_b}, \sigma_{t_a, t_b}) \\
\text{where:} \ &\mu_{t_a, t_b} := s(t)x_{t_b} + (1 - s(t))x_{t_a}\\
&\sigma_{t_a, t_b} := s(t)(1 - s(t))\tau (t_b - t_a)\boldsymbol{I} \\
&s(t) := (t - t_a)/(t_b - t_a)
\end{split}
\end{equation}

Eq.~\ref{eq:SBsim1} provides a framework for simulating $T^{\star}$. Instead of tracking $T^{\star}_{t_a, t_b}$ for a randomly selected time interval, we fix the terminal conditional and let the states evolve from beginning to end.  Specifically, let $\mathbf{t} = \{t_i\}_{i=0}^N$ represent a set of predefined time steps and fix $t_b = 1$. As $t_a$ iterates over the elements of $\mathbf{t}$, $T^{\star}_{t_a,1}$ always represents the optimal joint marginal distribution at each step once Eq.~\ref{eq:SBsim1} solved, since its solution will maintain the marginal distribution of $T^{\star}$ over the time interval we selected. Furthermore, for each $t_i \in \mathbf{t}$, the joint distribution $p(x_1,x_{t_i})$ (or its sample since we need to simulate its distribution) required by Eq.~\ref{eq:SBsim1} can be obtained by the following decompositions:
\begin{equation}\label{eq:SBsim3}
\begin{split}
    &p(x_1,x_{t_i}) := p(x_1|x_{t_i})p(x_{t_i}).\\
    &p(x_{t_i}) := p(x_0)p(x_1|x_0)\cdots p(x_{t_i}|x_{t_{i-1}})\\
    &\text{where:} \quad p(x_{t_{j+1}} | x_{t_j}) := \mathbb{E}_{p(x_1 | x_{t_j})}[p(x_{t_{j+1}} | x_{1}, x_{t_j})]\quad \text{and} \quad x_0\sim p_0
\end{split}
\end{equation}
\noindent Here, the Markovian property comes from Eq.~\ref{eq:sb-dynamic} (i.e., the Wiener measure is selected as reference measure), and distribution $p(x_{t_{j+1}} | x_{1}, x_{t_j})$ can be simulated with Eq.~\ref{eq:SBsim2} under a predefined time step. Consequently, once we know how to approximate the distribution $p(x_1 | x_{t_j})$ for all $j=0,1,\dots,i-1$, Eq.~\ref{eq:SBsim3} can be used to approximate the desired distributions $p(x_1,x_{t_i})$ iteratively as Fig.~\ref{fig:sb-model-ill}.

\subsubsection{Unpaired Neural Bridge Learning.}

\begin{figure}[t]
    \centering
    \includegraphics[width=\linewidth]{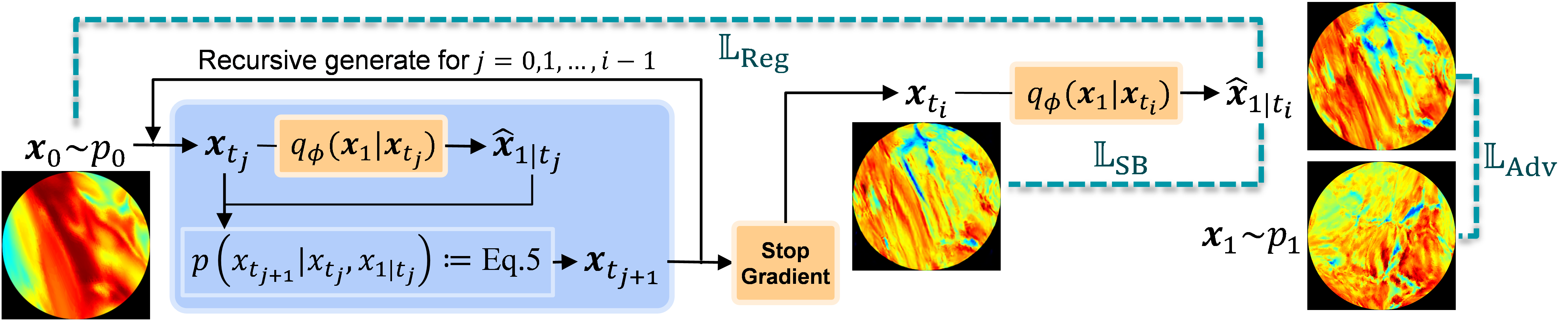}
    \caption{The illustration of training pipeline. For a randomly selected time step $t_i\in \mathbf{t}$, we recursively generate samples following Eq.~\ref{eq:SBsim2} and Eq.~\ref{eq:SBsim3} to approximate distribution $\hat{x}_{1 \mid t_i} \sim p(x_1,x_{t_i})$ as discussed in Sec.~\ref{sec:Method-sb}. All three losses $\mathbb{L}_\text{Adv}, \mathbb{L}_\text{SB}$ and $\mathbb{L}_\text{reg}$ are weighted through individual hyper-parameters.}
    \label{fig:sb-model-ill}
\end{figure}

From previous discussion, we know that it is necessary to obtain the posterior distribution $p(x_1 \mid x_{t_i})$ to compute the objective function in Eq.~\ref{eq:SBsim1} over an arbitrary sub-interval $[t_i, 1]$. To achieve this, we employ a neural generator parameterized by $\phi$, denoted as $q_\phi(x_1 \mid x_{t_i})$. Since the network needs to identity the time-dependent distributions, the generator takes $x_{t_i}$ and $t_i$ as inputs. Consequently, the objective function in  Eq.~\ref{eq:SBsim1} over the sub-interval $[t_i, 1]$ can be reformulated as:
\begin{equation}
\begin{split}
&\min_\phi \mathbb{L}_{\text{SB}}(\phi, t_i) := \mathbb{E}_{q_\phi(x_{t_i}, x_1)}\|x_{t_i} - x_1\|^2 - 2\tau (1 - t_i) H(q_\phi(x_{t_i}, x_1))
\\
&\text{subject to:} \quad \mathbb{L}_{\text{Adv}}(\phi, t_i) := D_{\text{KL}}(q_\phi(x_1) \parallel p(x_1)) = 0
\end{split}
\label{eq:loss}
\end{equation}

\noindent where $q_\phi(x_{t_i}, x_1) := q_\phi(x_1 \mid x_{t_i})p(x_{t_i})$, and the constraint ensures that the generator learns the high-quality fMRI distribution. By introducing a Lagrange multiplier, Eq.~\ref{eq:loss} can be reformulated into the following loss function:
\begin{equation}
\min_\phi \mathbb{L}(\phi, t_i) := \mathbb{L}_{\text{Adv}}(\phi, t_i) + \lambda_{\text{SB}} \mathbb{L}_{\text{SB}}(\phi, t_i)
\label{eq:loss_lambda}
\end{equation}

By solving Eq.~\ref{eq:loss_lambda} with the optimal $\phi$, the generator achieves $ q_\phi(x_1 \mid x_{t_i}) =  p(x_1 \mid x_{t_i})$ and $q_\phi(x_{t_i}, x_1) = p(x_{t_i}, x_1)$ for every steps $t_i$~\cite{kim2023unpaired,dong2024cunsb}. Consequently, the generator $q_\phi(x_1 \mid x_{t_i})$ can be directly utilized to sample the next step enhanced fMRI response $x_{t_{i+1}}$ for every steps $i = 0,1,\dots,N-1$. Through the iterative process shown in Fig.~\ref{fig:sb-model-ill}, we can generate the final enhanced fMRI response $x_{t_N}$ (i.e. the predicted $\hat{x}_1$) starting from the initial 3T distribution $x_0 \sim p_0$. The training and inference pipeline follows the procedure outlined in Sec.~\ref{sec:training}.

However, optimizing Eq.~\ref{eq:loss_lambda} alone does not guarantee the enhanced fMRI $x_1$ to preserve the structural details of its low-quality counterpart $x_0$, since Eq.~\ref{eq:loss_lambda} alone only ensures the optimal transformation path. To address this limitation, PatchNCE~\cite{park2020contrastive,kim2023unpaired} and MS-Structural Similarity Measure (MS-SSIM)~\cite{dong2024cunsb} regularization are incorporated. Therefore, the final loss function is defined as:
\begin{equation}\label{eq:loss_final}
\mathbb{L}(\phi, t_i) := \mathbb{L}_{\text{Adv}}(\phi, t_i) + \lambda_{\text{SB}} \mathbb{L}_{\text{SB}}(\phi, t_i) + \sum_l \lambda_{\text{Reg}_l} \mathbb{L}_{\text{Reg}_l}(\phi, t_i)
\end{equation}
\noindent where $\lambda_{\text{SB}}$ and $\lambda_{\text{Reg}_l}$ represent weights of SB loss and regularization terms.

\subsection{Re-Sampling and pRF Analysis}\label{sec:Method-prf}

With a well-trained Schrödinger Bridge diffusion model, we can generate enhanced versions of low-quality fMRI BDs. Due to the bijective nature of the conformal mapping for BDs, the fMRI response for every vertex can be accurately re-sampled from the enhanced BDs. By aggregating the enhanced fMRI responses across all parametrized time slices within a single pRF session, we reconstruct the complete enhanced fMRI time series.

To evaluate the effectiveness of our model, we employ pRF decoding as a downstream task, leveraging the enhanced fMRI time series to quantify improvements in receptive field estimation. Given a voxel-wise (or vertex-wise for surface based analysis) fMRI time series signal $\boldsymbol{y} = \{y_j(t)\}$ from a visual field stimuli experiment, the pRF model~\cite{dumoulin2008population,kay2013compressive} predicts the receptive center $ \boldsymbol{v}_j = (v_j^{(1)}, v_j^{(2)}) $ and size $ \sigma_j $ of each vertex $j$ on the visual field. The predicted fMRI signal for vertex $j$ is given by:
\begin{equation}
\hat{y}_j(\boldsymbol{v}_j, \sigma_j, t) = \beta \left[ \int r(\boldsymbol{x}; \boldsymbol{v}_j, \sigma_j) s(t, \boldsymbol{x}) \, d\boldsymbol{x} \right] * h(t)
\end{equation}
where $ \beta $ is a scaling coefficient, $ h(t) $ is the hemodynamic response function, and $ r(\boldsymbol{x}; \boldsymbol{v}, \sigma) $ is a Gaussian kernel $ \propto \exp\left(-\frac{(v^{(1)} - x^{(1)})^2 + (v^{(2)} - x^{(2)})^2}{2\sigma^2}\right) $. The parameters $ (\boldsymbol{v}_j, \sigma_j) $ are estimated by minimizing the prediction error:
\begin{equation}\label{eq:prf}
(\boldsymbol{v}_j, \sigma_j) = \arg \min_{\boldsymbol{v}_j, \sigma_j} \sum_t \| \hat{y}_j(\boldsymbol{v}, \sigma, t) - y_j(t) \|^2
\end{equation}

The pRF results of the entire visual cortex are obtained by solving Eq.~\ref{eq:prf} for every vertices on the cortical surface. The quality of fit is usually evaluated using variance explained $R_j^2$ for each vertex $j$ (in percentage):
\begin{equation}
 R_j^2 := (1 - \frac{ \sum_t (\hat{y}_j(t) - y_j(t))^2}{\sum_t (y_j(t) - \bar{y}_j)^2}    )\times100 \%
\end{equation}

\section{Experiments and Results}

\subsection{Experimental Details}

\subsubsection{Synthetic Data Down-sampling.}\label{results:downsample}
To simulate low-quality fMRI time series from the original 7T NSD data, we use Neuromaps~\cite{markello2022neuromaps} to transform all pRF sessions from the 164k fsaverage surface to the 32k fsLR surface format, matching the resolution of the 3T NOD data. Gaussian noise with $0$ mean and $5.0$ standard deviation is then added to each vertex in the transformed data to simulate signal degradation. This process creates a collection of synthetic low-quality data with corresponding high-quality counterparts, enabling ground-truth evaluations.

\subsubsection{Training Settings.}\label{sec:training}
The generator and discriminator architectures follow the designs outlined in~\cite{kim2023unpaired,dong2024cunsb}, and we set the number of time-steps $N = 5$. The loss weights in Eq.~\ref{eq:loss_final} are set as follows: $\lambda_{SB} = 1$, $\lambda_{\text{Reg}_1} = 0.5$, and $\lambda_{\text{Reg}_2} = 1$, corresponding to the PatchNCE~\cite{park2020contrastive,kim2023unpaired} and MS-SSIM~\cite{brunet2011mathematical,dong2024cunsb} regularization losses.

The model is trained for 150 epochs using the Adam optimizer, with an initial learning rate of $1 \times 10^{-4}$ that decays linearly to zero after the first 75 epochs. All input brain disks are resized to $256 \times 256$ without data augmentation. The batch size is set to 8. Training is conducted on a single \textit{NVIDIA GeForce GTX TITAN X}, requiring approximately 45 GPU hours.

\subsubsection{Evaluation.} 
For real 3T data without available high-quality ground truth, we evaluate the enhancement using the Fréchet Inception Distance (FID)~\cite{heusel2017gans}, which measures the general dissimilarity between the high-quality 7T fMRI collection, the original 3T fMRI, and the enhanced 3T fMRI.

For synthetic data, where ground truth is available, we include the structural similarity index measure (SSIM) and the peak signal-to-noise ratio (PSNR) as additional metrics. These scores are computed between the enhanced 3T fMRI signals and their corresponding high-quality 7T ground truth.

As a downstream task, the enhanced fMRI data undergoes pRF analysis using the analyzepRF code package~\cite{kay2013compressive}, with validation performed through \textit{qpRF}, a faster implementation~\cite{waz2024qprf}. Furthermore, we compare the original BOLD time series signal with the predicted signal generated using standard pRF parameters and the enhanced fMRI parameters, respectively.

\subsection{Results}

\renewcommand{\arraystretch}{1.2}
\setlength{\tabcolsep}{4pt}
\begin{table}[t!]
\centering
\caption{Performance for enhanced fMRI response and pRF results. For real data, we show the FID and the average $R^2$ value across all ROI vertices. For synthetic data in which we have paired ground truth, we use the average SSIM and PSNR value across all fMRI time series as additional metrics. Our enhancement achieved higher SSIM, PSNR and $R^2$, while maintaining a lower FID score.}
\resizebox{0.8\textwidth}{!}{%
\begin{tabular}{c|cc|ccc}
\toprule
\multirow{2}{*}{ Metrics }  & \multicolumn{2}{c|}{Real data} & \multicolumn{3}{c}{Synthetic data} \\ \cline{2-6}
 & Original & Enhanced & Down-Sampled & Enhanced & GT \\ \hline
SSIM $\uparrow$ & - & - & $0.7467$ & $\mathbf{0.8584}$ & - \\
PSNR $\uparrow$ & - & - & $20.5299$ & $\mathbf{24.8780}$ & - \\
FID $\downarrow$ & $183.8332$ & $\mathbf{70.6537}$ & $69.9228$ & $\mathbf{42.6448}$ & - \\ \hline
$\bar{R^2}$ $\uparrow$ & $25.9085$ & $\mathbf{28.5358}$ & $15.2975$ & $\mathbf{24.0026}$ & $24.0057$ \\  \bottomrule
\end{tabular}
}
\label{tab:metric}
\end{table}

\subsubsection{Enhanced fMRI Results.}

\begin{figure}[t!]
    \centering
    \includegraphics[width=0.85\linewidth]{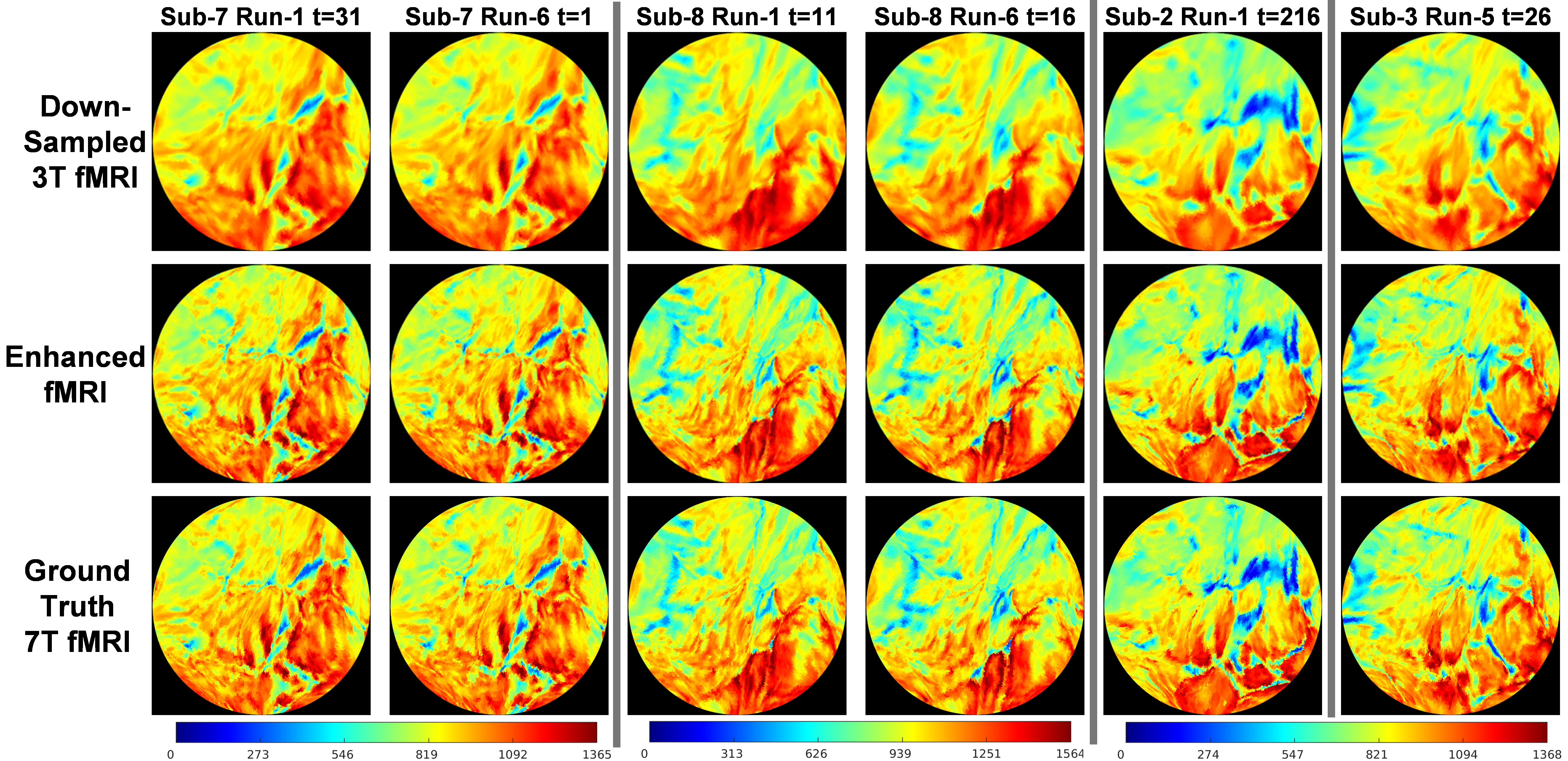}
    \caption{Illustration of enhanced BDs. We show the fMRI response on the shared parameterized planar disk.}
    \label{fig:enhanceBD}
\end{figure}

\begin{figure}[t!]
    \centering
    \includegraphics[width=0.8\linewidth]{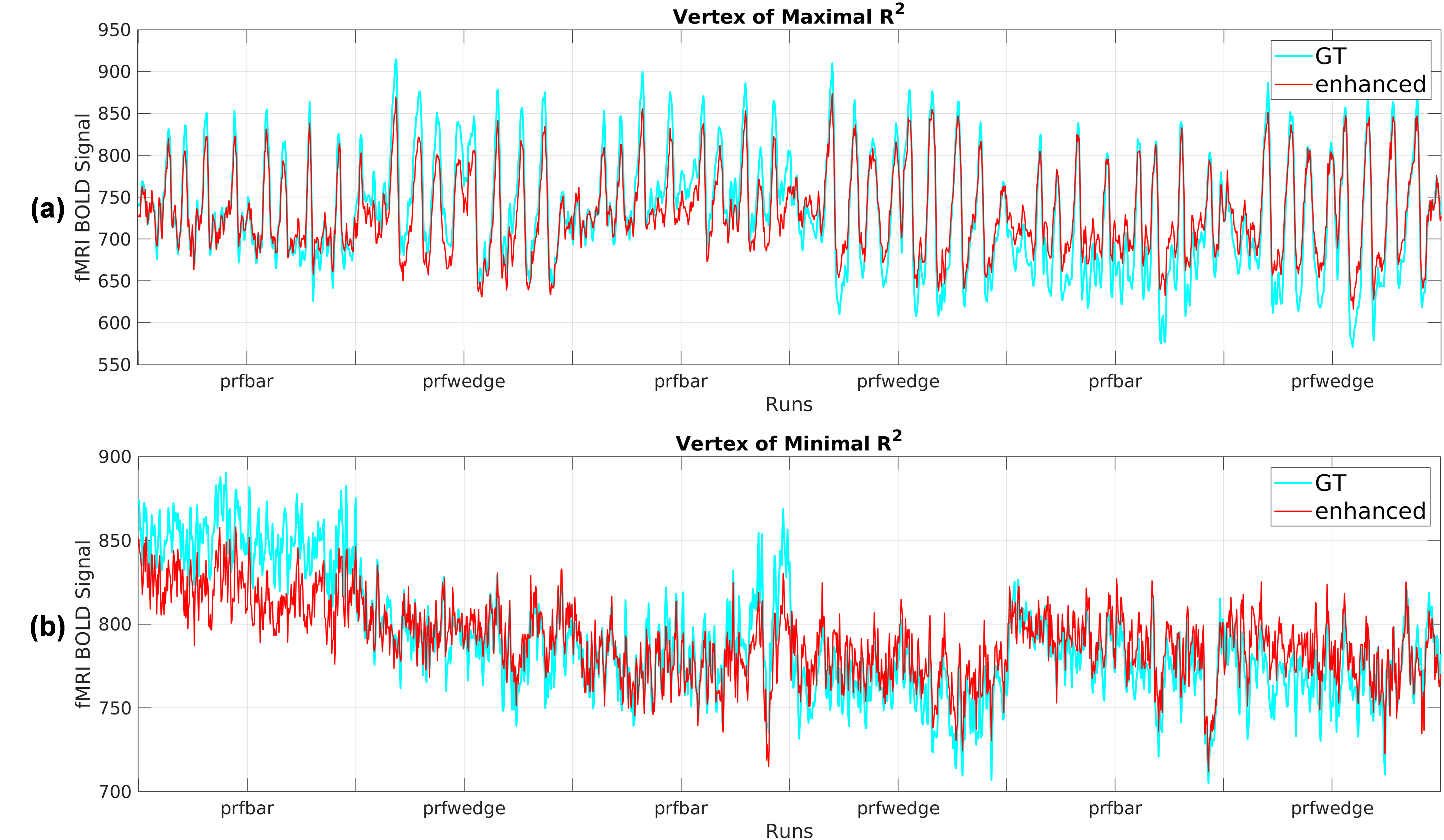}
    \caption{BOLD signal from synthetic low-quality data, enhanced data and the ground truth. (a) fMRI plots for the vertex with highest $R^2$ show a significant alignment between the ground truth and the enhanced signals, with slight misalignment on valley points. (b) fMRI plots for the vertex with lower $R^2$ show worse alignment, probably caused by the inactive and small visual response on this vertex.}
    \label{fig:bold}
\end{figure}

We evaluated enhanced fMRI on the same parametric brain disks obtained by conformal mapping, as illustrated in Fig.~\ref{fig:enhanceBD}. The enhanced disks exhibit finer resolution and a more distinct fMRI distribution that aligns closely with the underlying cortical structures, particularly in regions with extreme values or high curvature. To further analyze the performance, we compared the ground truth and the enhanced BOLD time series for two distinct vertices within our ROI: one with a strong response to the pRF stimuli and another with minimal response (determined by the value $R^2$ in the pRF analysis). As shown in Fig.~\ref{fig:bold}, the enhanced signal closely matches the ground truth for the active vertex, demonstrating a strong performance in capturing extreme values. However, for inert vertices, where the signal remains relatively constant, the alignment is weaker. This discrepancy likely arises from the difficulty in learning unchanging values based on their distribution across the brain disks.

Quantitative results are summarized in Tab.~\ref{tab:metric}. For real data, we achieved a $61.57\%$ reduction in the FID score, bringing the enhanced fMRI data distribution significantly closer to that of high-quality 7T fMRI collections while maintaining its brain structure. This improvement highlights the ability of our enhancement to achieve a level of precision on the fsaverage surface that is unattainable with the original 3T fMRI data on their lower-resolution surfaces. For synthetic data, where paired ground truth is available, we additionally evaluated the SSIM and PSNR between the down-sampled or enhanced fMRI brain disks and their ground truth counterparts. The enhancement results in a $14.96\%$ increase in SSIM and a $21.18\%$ increase in PSNR, effectively retrieving fMRI signals that closely approximate the ground truth and significantly reducing synthetic noise.

\subsubsection{Enhanced pRF Results.}
\begin{figure}[t!]
    \centering
    \includegraphics[width=0.7\linewidth]{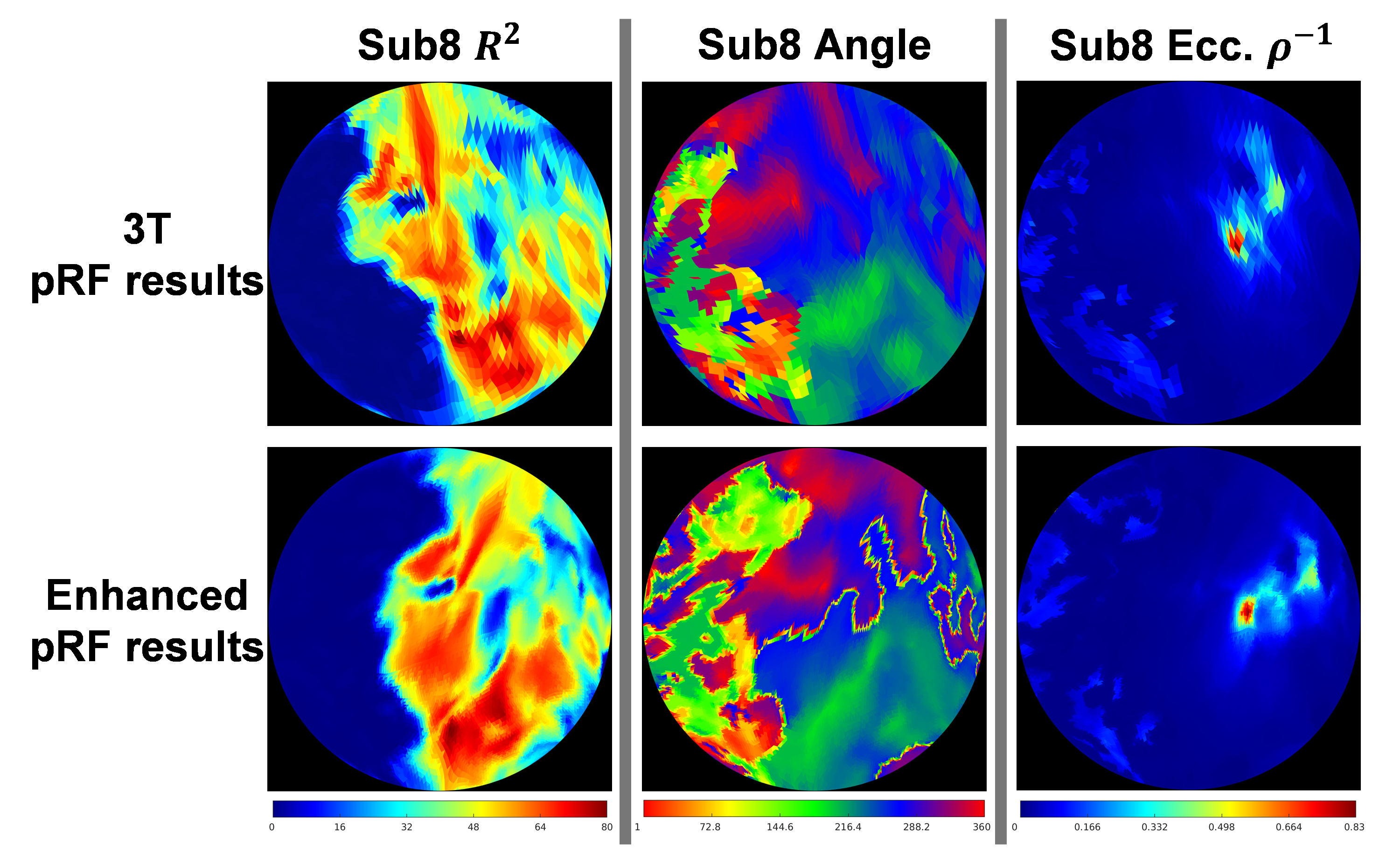}
    \caption{Illustration of pRF results from NOD test subject $s_8$. Our enhancement maintain the brain structure through their receptive fields (different vertex label from fsLR and fsaverage may cause ROI shifting), while achieved higher $R^2$ across the ROI.}
    \label{fig:enhancePRF}
\end{figure}

Beyond evaluating the enhanced fMRI signals directly, we perform downstream analysis to demonstrate the broader applicability of our pipeline. In Fig.~\ref{fig:enhancePRF}, we compare the original pRF results from the NOD dataset with the pRF results derived from enhanced fMRI time series, projected onto the parametric domain. The results show a notable improvement in the overall $R^2$ values across many vertices, while preserving the brain structure through their receptive fields, despite the absence of paired training.

\begin{figure}[t!]
    \centering
    \includegraphics[width=0.7\linewidth]{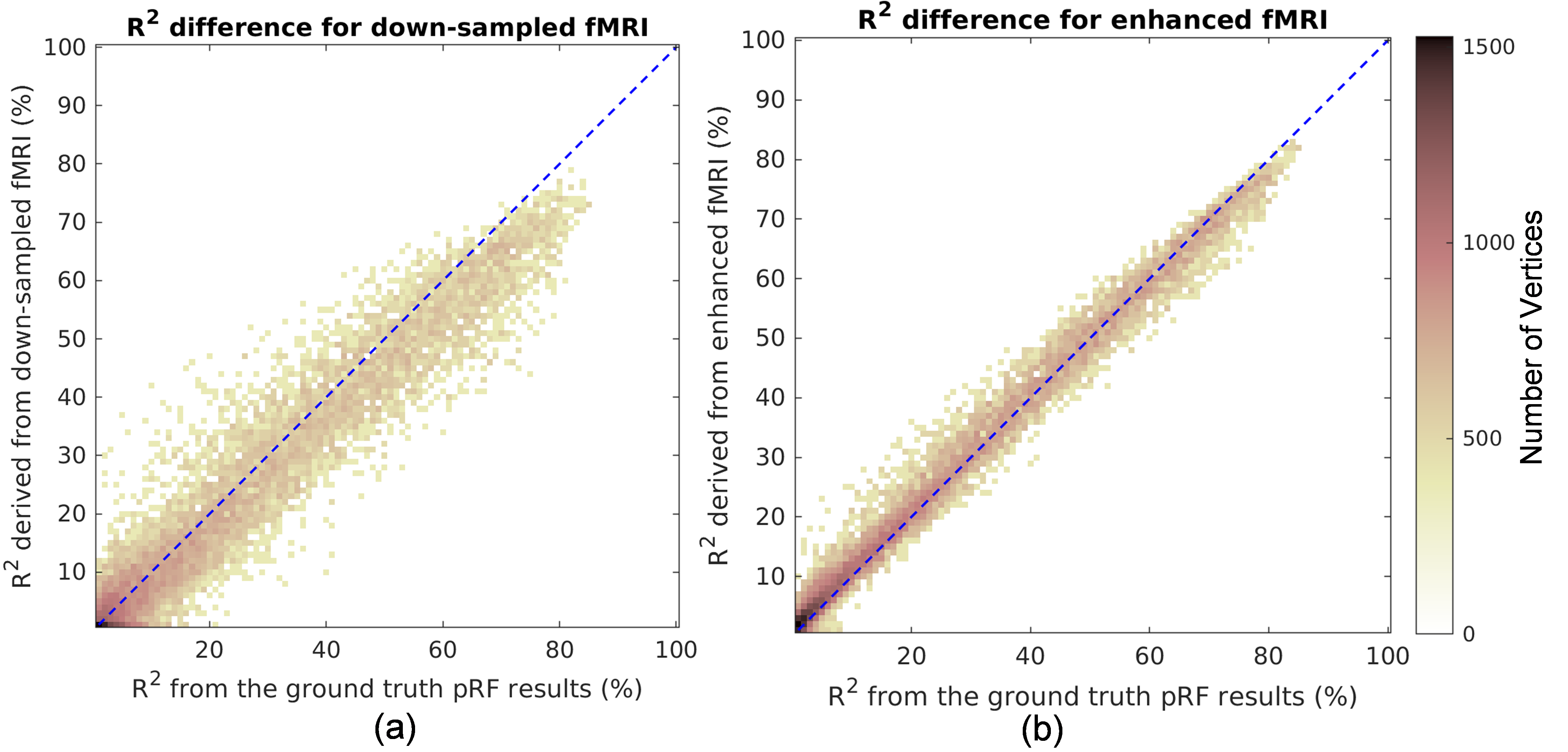}
    \caption{Heatmap comparing $R^2$ values. (a) The $R^2$ derived from down-sampled fMRI show a larger variation and generally lower than the $R^2$ from GT fMRI. (b) However, the $R^2$ derived from enhanced fMRI show a significant improvement in variance (especially for high $R^2$ vertices), and maintain similar confidential threshold to GT.}
    \label{fig:heatmap}
\end{figure}

For synthetic data, we further validate the performance by comparing the $R^2$ values of the native 7T pRF parameters with those derived from the enhanced fMRI data, as shown in Fig.~\ref{fig:heatmap}. The down-sampled fMRI data exhibit greater variance and generally lower $R^2$ values compared to the ground truth high-quality pRF results. In contrast, the $R^2$ values derived from the enhanced fMRI data show significant improvement, particularly for vertices with high $R^2$, while maintaining a comparable confidence threshold to the ground truth. These results confirm that our enhancement pipeline preserves the interpretability and reliability of pRF solutions, bridging the gap to 7T data in downstream tasks.

\section{Conclusion and Future Work}
We present a robust fMRI processing pipeline that transforms BOLD signals from 3D cortical surfaces to 2D parametric brain disks. We introduce an unsupervised Schrödinger Bridge diffusion model to enhance 3T brain disks using unpaired 7T data, demonstrating that the enhanced 3T brain disks achieve quality and downstream performance comparable to native 7T fMRI. Our method effectively preserves both visual information and cortical structure, crucial for tasks like pRF modeling. By matching the performance of high-quality 7T data, our approach paves the way for advanced visual fMRI analysis with 3T scans.

In addition to pRF analyses, our pipeline is capable of other downstream applications. With the enhanced fMRI signals for 3T experiments, we can derive a continuous model for segmentation and classification tasks. Our future work will generalize the SB enhancement pipeline to process raw 3D fMRI data.
In general, our framework is promising to become the leading method to enhance the quality of common 3T fMRI to 7T comparable for various applications.


%
%
%
\newpage

%
%
%
\bibliographystyle{splncs04}
\bibliography{refs}

\begin{thebibliography}{10}
\providecommand{\url}[1]{\texttt{#1}}
\providecommand{\urlprefix}{URL }
\providecommand{\doi}[1]{https://doi.org/#1}

\bibitem{allen2022massive}
Allen, E.J., St-Yves, G., Wu, Y., Breedlove, J.L., Prince, J.S., Dowdle, L.T., Nau, M., Caron, B., Pestilli, F., Charest, I., et~al.: A massive 7t fmri dataset to bridge cognitive neuroscience and artificial intelligence. Nature neuroscience  \textbf{25}(1),  116--126 (2022)

\bibitem{armanious2019unsupervised}
Armanious, K., Jiang, C., Abdulatif, S., K{\"u}stner, T., Gatidis, S., Yang, B.: Unsupervised medical image translation using cycle-medgan. In: 2019 27th European signal processing conference (EUSIPCO). pp.~1--5. IEEE (2019)

\bibitem{benson2018human}
Benson, N.C., Jamison, K.W., Arcaro, M.J., Vu, A.T., Glasser, M.F., Coalson, T.S., Van~Essen, D.C., Yacoub, E., Ugurbil, K., Winawer, J., et~al.: The human connectome project 7 tesla retinotopy dataset: Description and population receptive field analysis. Journal of vision  \textbf{18}(13),  23--23 (2018)

\bibitem{benson2022variability}
Benson, N.C., Yoon, J.M., Forenzo, D., Engel, S.A., Kay, K.N., Winawer, J.: Variability of the surface area of the v1, v2, and v3 maps in a large sample of human observers. Journal of Neuroscience  \textbf{42}(46),  8629--8646 (2022)

\bibitem{brewer2014visual}
Brewer, A.A., Barton, B.: Visual cortex in aging and alzheimer's disease: changes in visual field maps and population receptive fields. Frontiers in psychology  \textbf{5}, ~74 (2014)

\bibitem{brunet2011mathematical}
Brunet, D., Vrscay, E.R., Wang, Z.: On the mathematical properties of the structural similarity index. IEEE Transactions on Image Processing  \textbf{21}(4),  1488--1499 (2011)

\bibitem{chang2019bold5000}
Chang, N., Pyles, J.A., Marcus, A., Gupta, A., Tarr, M.J., Aminoff, E.M.: Bold5000, a public fmri dataset while viewing 5000 visual images. Scientific data  \textbf{6}(1), ~49 (2019)

\bibitem{chen2024contourdiff}
Chen, Y., Konz, N., Gu, H., Dong, H., Chen, Y., Li, L., Lee, J., Mazurowski, M.A.: Contourdiff: Unpaired image translation with contour-guided diffusion models. arXiv preprint arXiv:2403.10786  (2024)

\bibitem{chen2024cinematic}
Chen, Z., Qing, J., Zhou, J.H.: Cinematic mindscapes: High-quality video reconstruction from brain activity. Advances in Neural Information Processing Systems  \textbf{36} (2024)

\bibitem{cui20247t}
Cui, Q., Tosun, D., Mukherjee, P., Abbasi-Asl, R.: 7t mri synthesization from 3t acquisitions. In: International Conference on Medical Image Computing and Computer-Assisted Intervention. pp. 35--44. Springer (2024)

\bibitem{deyoe1996mapping}
DeYoe, E.A., Carman, G.J., Bandettini, P., Glickman, S., Wieser, J., Cox, R., Miller, D., Neitz, J.: Mapping striate and extrastriate visual areas in human cerebral cortex. Proceedings of the National Academy of Sciences  \textbf{93}(6),  2382--2386 (1996)

\bibitem{dickie2019ciftify}
Dickie, E.W., Anticevic, A., Smith, D.E., Coalson, T.S., Manogaran, M., Calarco, N., Viviano, J.D., Glasser, M.F., Van~Essen, D.C., Voineskos, A.N.: Ciftify: A framework for surface-based analysis of legacy mr acquisitions. Neuroimage  \textbf{197},  818--826 (2019)

\bibitem{dong2024cunsb}
Dong, X., Vasa, V.K., Zhu, W., Qiu, P., Chen, X., Su, Y., Xiong, Y., Yang, Z., Chen, Y., Wang, Y.: {CUNSB-RFIE}: Context-aware unpaired neural schrödinger bridge in retinal fundus image enhancement. arXiv preprint arXiv:2409.10966  (2024)

\bibitem{dougherty2003visual}
Dougherty, R.F., Koch, V.M., Brewer, A.A., Fischer, B., Modersitzki, J., Wandell, B.A.: Visual field representations and locations of visual areas v1/2/3 in human visual cortex. Journal of vision  \textbf{3}(10), ~1--1 (2003)

\bibitem{dumoulin2003automatic}
Dumoulin, S.O., Hoge, R.D., Baker~Jr, C.L., Hess, R.F., Achtman, R.L., Evans, A.C.: Automatic volumetric segmentation of human visual retinotopic cortex. Neuroimage  \textbf{18}(3),  576--587 (2003)

\bibitem{dumoulin2008population}
Dumoulin, S.O., Wandell, B.A.: Population receptive field estimates in human visual cortex. Neuroimage  \textbf{39}(2),  647--660 (2008)

\bibitem{duncan2007retinotopic}
Duncan, R.O., Sample, P.A., Weinreb, R.N., Bowd, C., Zangwill, L.M.: Retinotopic organization of primary visual cortex in glaucoma: Comparing fmri measurements of cortical function with visual field loss. Progress in retinal and eye research  \textbf{26}(1),  38--56 (2007)

\bibitem{engel1994fmri}
Engel, S.A., Rumelhart, D.E., Wandell, B.A., Lee, A.T., Glover, G.H., Chichilnisky, E.J., Shadlen, M.N., et~al.: fmri of human visual cortex. Nature  \textbf{369}(6481),  525--525 (1994)

\bibitem{fang2020reconstructing}
Fang, T., Qi, Y., Pan, G.: Reconstructing perceptive images from brain activity by shape-semantic gan. Advances in Neural Information Processing Systems  \textbf{33},  13038--13048 (2020)

\bibitem{fischl2012freesurfer}
Fischl, B.: Freesurfer. Neuroimage  \textbf{62}(2),  774--781 (2012)

\bibitem{glasserSupplementaryMaterialMultimodal2016}
Glasser, M.F., Coalson, T.S., Robinson, E.C., Hacker, C.D., Harwell, J., Yacoub, E., Ugurbil, K., Andersson, J., Beckmann, C.F., Jenkinson, M., Smith, S.M., Van~Essen, D.C.: Supplementary {{Material}}: {{A}} multi-modal parcellation of human cerebral cortex. Nature  \textbf{536}(7615),  171--178 (2016). \doi{10.1038/nature18933}, \url{https://www.nature.com/articles/nature18933}

\bibitem{glasser2016multi}
Glasser, M.F., Coalson, T.S., Robinson, E.C., Hacker, C.D., Harwell, J., Yacoub, E., Ugurbil, K., Andersson, J., Beckmann, C.F., Jenkinson, M., et~al.: A multi-modal parcellation of human cerebral cortex. Nature  \textbf{536}(7615),  171--178 (2016)

\bibitem{gong2023large}
Gong, Z., Zhou, M., Dai, Y., Wen, Y., Liu, Y., Zhen, Z.: A large-scale fmri dataset for the visual processing of naturalistic scenes. Scientific Data  \textbf{10}(1), ~559 (2023)

\bibitem{goodfellow2020generative}
Goodfellow, I., Pouget-Abadie, J., Mirza, M., Xu, B., Warde-Farley, D., Ozair, S., Courville, A., Bengio, Y.: Generative adversarial networks. Communications of the ACM  \textbf{63}(11),  139--144 (2020)

\bibitem{gu2004genus}
Gu, X., Wang, Y., Chan, T.F., Thompson, P.M., Yau, S.T.: Genus zero surface conformal mapping and its application to brain surface mapping. IEEE transactions on medical imaging  \textbf{23}(8),  949--958 (2004)

\bibitem{heusel2017gans}
Heusel, M., Ramsauer, H., Unterthiner, T., Nessler, B., Hochreiter, S.: Gans trained by a two time-scale update rule converge to a local nash equilibrium. Advances in neural information processing systems  \textbf{30} (2017)

\bibitem{himmelberg2023comparing}
Himmelberg, M.M., T{\"u}n{\c{c}}ok, E., Gomez, J., Grill-Spector, K., Carrasco, M., Winawer, J.: Comparing retinotopic maps of children and adults reveals a late-stage change in how v1 samples the visual field. Nature communications  \textbf{14}(1), ~1561 (2023)

\bibitem{ho2020denoising}
Ho, J., Jain, A., Abbeel, P.: Denoising diffusion probabilistic models. Advances in neural information processing systems  \textbf{33},  6840--6851 (2020)

\bibitem{horikawa2017generic}
Horikawa, T., Kamitani, Y.: Generic decoding of seen and imagined objects using hierarchical visual features. Nature communications  \textbf{8}(1),  15037 (2017)

\bibitem{huang2024noise}
Huang, S., Luo, G., Wang, X., Chen, Z., Wang, Y., Yang, H., Heng, P.A., Zhang, L., Lyu, M.: Noise level adaptive diffusion model for robust reconstruction of accelerated mri. In: International Conference on Medical Image Computing and Computer-Assisted Intervention. pp. 498--508. Springer (2024)

\bibitem{jin2018conformal}
Jin, M., Gu, X., He, Y., Wang, Y.: Conformal geometry. Computational Algorithms  (2018)

\bibitem{kay2013compressive}
Kay, K.N., Winawer, J., Mezer, A., Wandell, B.A.: Compressive spatial summation in human visual cortex. Journal of neurophysiology  \textbf{110}(2),  481--494 (2013)

\bibitem{kim2023unpaired}
Kim, B., Kwon, G., Kim, K., Ye, J.C.: Unpaired image-to-image translation via neural schrödinger bridge. arXiv preprint arXiv:2305.15086  (2023)

\bibitem{konz2024anatomically}
Konz, N., Chen, Y., Dong, H., Mazurowski, M.A.: Anatomically-controllable medical image generation with segmentation-guided diffusion models. In: International Conference on Medical Image Computing and Computer-Assisted Intervention. pp. 88--98. Springer (2024)

\bibitem{korotin2023light}
Korotin, A., Gushchin, N., Burnaev, E.: Light schrödinger bridge. arXiv preprint arXiv:2310.01174  (2023)

\bibitem{leonard2013survey}
L{\'e}onard, C.: A survey of the schr$\backslash$" odinger problem and some of its connections with optimal transport. arXiv preprint arXiv:1308.0215  (2013)

\bibitem{li2020magnetic}
Li, W., Li, Y., Qin, W., Liang, X., Xu, J., Xiong, J., Xie, Y.: Magnetic resonance image (mri) synthesis from brain computed tomography (ct) images based on deep learning methods for magnetic resonance (mr)-guided radiotherapy. Quantitative imaging in medicine and surgery  \textbf{10}(6), ~1223 (2020)

\bibitem{markello2022neuromaps}
Markello, R.D., Hansen, J.Y., Liu, Z.Q., Bazinet, V., Shafiei, G., Su{\'a}rez, L.E., Blostein, N., Seidlitz, J., Baillet, S., Satterthwaite, T.D., et~al.: Neuromaps: structural and functional interpretation of brain maps. Nature Methods  \textbf{19}(11),  1472--1479 (2022)

\bibitem{park2020contrastive}
Park, T., Efros, A.A., Zhang, R., Zhu, J.Y.: Contrastive learning for unpaired image-to-image translation. In: Computer Vision--ECCV 2020: 16th European Conference, Glasgow, UK, August 23--28, 2020, Proceedings, Part IX 16. pp. 319--345. Springer (2020)

\bibitem{phan2023structure}
Phan, V.M.H., Liao, Z., Verjans, J.W., To, M.S.: Structure-preserving synthesis: Maskgan for unpaired mr-ct translation. In: International Conference on Medical Image Computing and Computer-Assisted Intervention. pp. 56--65. Springer (2023)

\bibitem{ren2021reconstructing}
Ren, Z., Li, J., Xue, X., Li, X., Yang, F., Jiao, Z., Gao, X.: Reconstructing seen image from brain activity by visually-guided cognitive representation and adversarial learning. NeuroImage  \textbf{228},  117602 (2021)

\bibitem{ribeiro2024human}
Ribeiro, F.L., Benson, N.C., Puckett, A.M.: Human retinotopic mapping: from empirical to computational models of retinotopy  (2024)

\bibitem{sasaki2021unit}
Sasaki, H., Willcocks, C.G., Breckon, T.P.: Unit-ddpm: Unpaired image translation with denoising diffusion probabilistic models. arXiv preprint arXiv:2104.05358  (2021)

\bibitem{shen2020modeling}
Shen, Z., Fu, H., Shen, J., Shao, L.: Modeling and enhancing low-quality retinal fundus images. IEEE transactions on medical imaging  \textbf{40}(3),  996--1006 (2020)

\bibitem{stigliani2015temporal}
Stigliani, A., Weiner, K.S., Grill-Spector, K.: Temporal processing capacity in high-level visual cortex is domain specific. Journal of Neuroscience  \textbf{35}(36),  12412--12424 (2015)

\bibitem{ta2022quantitative}
Ta, D., Tu, Y., Lu, Z.L., Wang, Y.: Quantitative characterization of the human retinotopic map based on quasiconformal mapping. Medical image analysis  \textbf{75},  102230 (2022)

\bibitem{takagi2023high}
Takagi, Y., Nishimoto, S.: High-resolution image reconstruction with latent diffusion models from human brain activity. In: Proceedings of the IEEE/CVF Conference on Computer Vision and Pattern Recognition. pp. 14453--14463 (2023)

\bibitem{thielen2019deeprf}
Thielen, J., G{\"u}{\c{c}}l{\"u}, U., G{\"u}{\c{c}}l{\"u}t{\"u}rk, Y., Ambrogioni, L., Bosch, S.E., van Gerven, M.A.: Deeprf: Ultrafast population receptive field mapping with deep learning. bioRxiv p. 732990 (2019)

\bibitem{tong2023conditional}
Tong, A., Malkin, N., Huguet, G., Zhang, Y., Rector-Brooks, J., Fatras, K., Wolf, G., Bengio, Y.: Conditional flow matching: Simulation-free dynamic optimal transport. arXiv preprint arXiv:2302.00482  \textbf{2}(3) (2023)

\bibitem{tu2021topology}
Tu, Y., Ta, D., Lu, Z.L., Wang, Y.: Topology-preserving smoothing of retinotopic maps. PLoS computational biology  \textbf{17}(8),  e1009216 (2021)

\bibitem{uugurbil2013pushing}
U{\u{g}}urbil, K., Xu, J., Auerbach, E.J., Moeller, S., Vu, A.T., Duarte-Carvajalino, J.M., Lenglet, C., Wu, X., Schmitter, S., Van~de Moortele, P.F., et~al.: Pushing spatial and temporal resolution for functional and diffusion mri in the human connectome project. Neuroimage  \textbf{80},  80--104 (2013)

\bibitem{van2013wu}
Van~Essen, D.C., Smith, S.M., Barch, D.M., Behrens, T.E., Yacoub, E., Ugurbil, K., Consortium, W.M.H., et~al.: The wu-minn human connectome project: an overview. Neuroimage  \textbf{80},  62--79 (2013)

\bibitem{wandell2007visual}
Wandell, B.A., Dumoulin, S.O., Brewer, A.A.: Visual field maps in human cortex. Neuron  \textbf{56}(2),  366--383 (2007)

\bibitem{wang2015probabilistic}
Wang, L., Mruczek, R.E., Arcaro, M.J., Kastner, S.: Probabilistic maps of visual topography in human cortex. Cerebral cortex  \textbf{25}(10),  3911--3931 (2015)

\bibitem{wang2007brain}
Wang, Y., Lui, L.M., Gu, X., Hayashi, K.M., Chan, T.F., Toga, A.W., Thompson, P.M., Yau, S.T.: Brain surface conformal parameterization using riemann surface structure. IEEE Transactions on Medical Imaging  \textbf{26}(6),  853--865 (2007)

\bibitem{wang2024implicit}
Wang, Y., Yoon, S., Jin, P., Tivnan, M., Chen, Z., Hu, R., Zhang, L., Chen, Z., Li, Q., Wu, D.: Implicit image-to-image schrodinger bridge for ct super-resolution and denoising. arXiv preprint arXiv:2403.06069  (2024)

\bibitem{waz2024qprf}
Waz, S., Wang, Y., Lu, Z.L.: q{PRF}: A system to accelerate population receptive field decoding. bioRxiv  (2024)

\bibitem{xiong2023characterizing}
Xiong, Y., Tu, Y., Lu, Z.l., Wang, Y.: Characterizing visual cortical magnification with topological smoothing and optimal transportation. In: Medical Imaging 2023: Image Processing. vol. 12464, pp. 464--472. SPIE (2023)

\bibitem{zhang2024phy}
Zhang, J., Yan, R., Perelli, A., Chen, X., Li, C.: Phy-diff: Physics-guided hourglass diffusion model for diffusion mri synthesis. In: International Conference on Medical Image Computing and Computer-Assisted Intervention. pp. 345--355. Springer (2024)

\bibitem{zhu2023otre}
Zhu, W., Qiu, P., Dumitrascu, O.M., Sobczak, J.M., Farazi, M., Yang, Z., Nandakumar, K., Wang, Y.: Otre: where optimal transport guided unpaired image-to-image translation meets regularization by enhancing. In: International Conference on Information Processing in Medical Imaging. pp. 415--427. Springer (2023)

\bibitem{zhu2024diffusion}
Zhu, X., Zhang, W., Li, Y., O’Donnell, L.J., Zhang, F.: When diffusion mri meets diffusion model: A novel deep generative model for diffusion mri generation. In: International Conference on Medical Image Computing and Computer-Assisted Intervention. pp. 530--540. Springer (2024)

\end{thebibliography}

\end{document}